%% file: acl2019.tex
\documentclass[11pt,a4paper]{article}
\usepackage[utf8x]{inputenc}
\usepackage[T1]{fontenc}
\usepackage{float}
\usepackage[hyperref]{acl2019}
\usepackage{times}
\usepackage{latexsym}
\usepackage{multirow}
\usepackage{graphicx}
\usepackage{amsmath}
\usepackage{comment}
\graphicspath{ {images/} }
\usepackage{url}

\aclfinalcopy 

\usepackage{flushend}

\title{The TALP-UPC System for the WMT Similar Language Task: Statistical vs Neural Machine Translation} 

\author{Magdalena Biesialska \hspace{7mm} Lluis Guardia  \hspace{7mm} Marta R. Costa-juss\`a \\
 TALP Research Center, Universitat Politècnica de Catalunya, 08034 Barcelona \\  \texttt{magdalena.biesialska@upc.edu lluis.guardia@alu-etsetb.upc.edu}\\ \texttt{marta.ruiz@upc.edu}}

\date{}

\begin{document}
\maketitle

\begin{abstract}
Although the problem of similar language translation has been an area of research interest for many years, yet it is still far from being solved. In this paper, we study the performance of two popular approaches: statistical and neural. We conclude that both methods yield similar results; however, the performance varies depending on the language pair. While the statistical approach outperforms the neural one by a difference of 6 BLEU points for the Spanish-Portuguese language pair, the proposed neural model surpasses the statistical one by a difference of 2 BLEU points for Czech-Polish. In the former case, the language similarity (based on perplexity) is much higher than in the latter case. Additionally, we report negative results for the system combination with back-translation.

Our TALP-UPC system submission won 1st place for Czech$\rightarrow$Polish and 2nd place for Spanish$\rightarrow$Portuguese in the official evaluation of the 1st WMT Similar Language Translation task.
\end{abstract}

\section{Introduction}
Much research work has been done on language translation in the past decades. Given recent success of various machine translation (MT) systems, it is not surprising that some could consider similar language translation an already solved task. However, there are still remaining challenges that need to be addressed, such as limited resources or out-of-domain. Apart from these well-known, standard problems, we have discovered other under-researched phenomena within the task of similar language translation. Specifically, there exist languages from the same linguistic family that have a high degree of difference in alphabets, as it is the case for Czech-Polish, which may pose a challenge for MT systems. 

Neural MT has achieved the best results in many tasks, outperforming former statistical MT (SMT) methods \cite{sennrich2016wmt}. However, there are tasks where previous statistical MT approaches are still competitive, such as unsupervised machine translation \cite{artetxe2018unsupervised-smt,lample-etal-2018-phrase}. Motivated by the close proximity between the languages at hand and limited resources, in this article we aimed to determine whether the neural or the statistical approach is a better one to solve the given problem.

We report our results in the 1st Similar Language Translation WMT task \cite{barrault-EtAl:2019:WMT}. In the official evaluation, our Czech$\rightarrow$Polish and Spanish$\rightarrow$Portuguese translation systems were ranked 1st and 2nd respectively. The main contributions of our work are the neural and statistical MT systems trained for similar languages, as well as the strategies for adding monolingual corpora in neural MT. Additionally, we report negative results on the system combination by using back-translation and Minimum Bayes Risk \cite{mbr} techniques.

\section{Background}
In this section, we provide a brief overview of statistical (Phrase-based) and Neural-based MT approaches as well as strategies for exploiting monolingual data.

\subsection{Phrase-based Approach}
\input{pbapproach}

\subsection{Neural-based Approach}
\label{sec:neural_approach}
\input{nmtapproach}

\paragraph{Adding Monolingual Data} Although our proposed statistical MT model incorporates monolingual corpora, the supervised neural MT approach is not capable to make use of such data. However, recent studies have reported notable improvements in the translation quality when monolingual corpora were added to the training corpora, either through back-translation \cite{sennrich:2016} or copied corpus \cite{currey-etal-2017-copied}. Encouraged by those results and given the similarity of languages at hand, we propose to exploit monolingual data by leveraging back-translation as well as by simply copying target-side monolingual corpus and use it together with the original parallel data.

\section{System Combination with Back-translation}
\input{systemComb}

\section{Experimental Framework}
In this section we describe the data sets, data preprocessing as well as training and evaluation details for the PB and neural MT systems and the system combination.

\input{data_preprocessing}
\input{experimental}

\section{Results}
\input{results}

\section{Discussion}
\input{discussion}


\section{Future Work}
\input{futurework}

\section*{Acknowledgments}
The authors want to thank Pablo Gamallo, Jos\'e Ramom Pichel Campos and I\~naki Alegria for sharing their valuable insights on their language distance studies.

This work is supported in part by the Spanish Ministerio de Econom\'ia y Competitividad, the European Regional  Development  Fund  and  the  Agencia  Estatal  de  Investigaci\'on,  through  the  postdoctoral  senior grant Ram\'on y Cajal, the contract TEC2015-69266-P (MINECO/FEDER,EU) and the contract PCIN-2017-079 (AEI/MINECO).

\bibliography{acl2019}
\bibliographystyle{acl_natbib}

\end{document}

%% file: pbapproach.tex
Phrase-based (PB) statistical MT \cite{koehn:2003} translates by concatenating at a phrase level the most probable target given the source text.
In this context, a phrase is a sequence of words, regardless if it is a phrase or not from the linguistic point of view. Phrases are extracted from word alignments between both languages in a large parallel corpus, based on the probabilistic study, which identifies each phrase with several features, such as conditional probabilities. The collection of scored phrases constitutes the translation model.

\begin{figure}[H]
\centering
\includegraphics[width=0.5\textwidth]{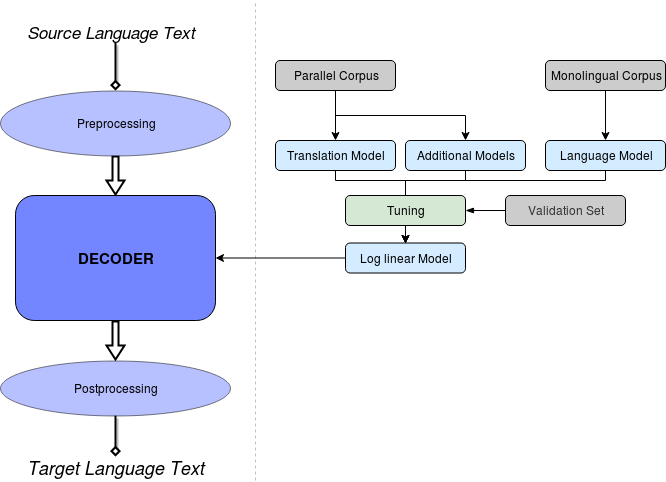}
\caption{\label{fig:smt}Basic schema of a Phrase-based MT system}
\end{figure}

In addition to this model, there are also other models to help achieve a better translation, such as the reordering model, which helps in a better ordering of the phrases; or the language model, trained from a monolingual corpus in the target language helping to obtain a better fluency in the translation. The weights of each of these models are optimized by tuning over a validation set. Based on these optimized combinations, the decoder uses beam search to find the most probable output given an input. Figure \ref{fig:smt} shows a diagram of the Phrase-based MT approach.

%% file: nmtapproach.tex
Neural networks (NNs) have been successful in many Natural Language Processing (NLP) tasks in recent years. NMT systems, which use end-to-end NN models to encode a source sequence in one language and decode a target sequence in the second language, early on demonstrated performance on a par with or even outperformed traditional Phrase-based SMT systems \cite{kalchbrenner2013,cho2014learning,sutskever2014,bahdanau2014neural,sennrich2016wmt,zhou2016,wu2016}.

Previous state-of-the-art NMT models used predominantly bi-directional recurrent neural networks (RNN) equipped with Long-Short Term Memory (LSTM; \citealt{hochreiter1997}) units or Gated Recurrent Units (GRU; \citealt{cho2014learning}) both in the encoder and the decoder combined with the attention mechanism \cite{bahdanau2014neural,luong2015}. There were also approaches, although less common, to leverage convolutional neural networks (CNN) for sequence modeling \cite{kalchbrenner2O16,gehring2017}.

In this work, we focus on the most current state-of-the-art NMT architecture, the Transformer \cite{vaswani2017attention}, which shows significant performance improvements over traditional sequence-to-sequence models. Interestingly, while the Transformer employs many concepts that were used earlier in encoder-decoder RNN and CNN based models, such as: residual connections \cite{he2015}, position embeddings \cite{gehring2017}, attention; the Transformer architecture relies solely on the self-attention mechanism without resorting to either recurrence or convolution.

The variant of the self-attention mechanism implemented by the Transformer, multi-head attention, allows to model dependencies between all tokens in a sequence irrespective of their actual position. More specifically, the representation of a given word is produced by means of computing a weighted average of attention scores of all words in a sentence.

%% file: systemComb.tex
\label{sec:backtranslation}

In this paper, we propose to combine the results of both Phrase-based and NMT systems at the sentence level. However, differently from the previous work of \newcite{marie-fujita-2018-smorgasbord}, we aimed for a conceptually simple combination strategy.

\begin{figure*}[!h]
\includegraphics[width=1\textwidth]{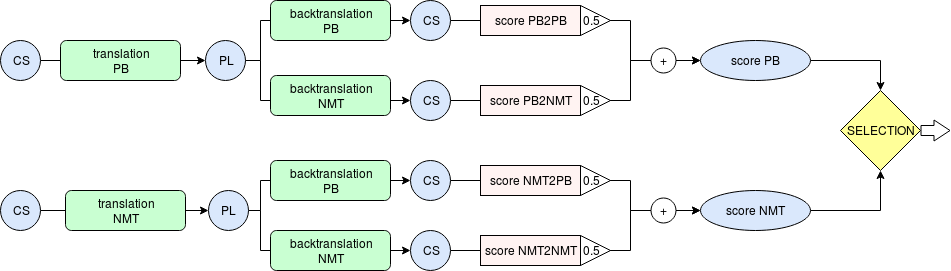}
\caption{\label{fig:combination}Scheme of the system combination approach}
\end{figure*}

In principle, for every sentence generated by the two alternative systems we used the BLEU score \cite{papineni2002bleu} to select a sentence with the highest translation quality. Each of the translations was back-translated (i.e. translated from the target language to the source language). Instead of using only one system to perform back-translation, we used both PB and neural MT systems and weighted them equally. See Figure \ref{fig:combination} for a graphical representation of this strategy.

This approach was motivated by the recent success of different uses of back-translation in neural MT studies \cite{sennrich:2016,lample-etal-2018-phrase}.
The final test set was composed of sentences produced by the system that obtained the highest score based on the quality of the combined back-translation. 

%% file: data_preprocessing.tex
\subsection{Data and Preprocessing}
Both submitted systems are constrained, hence they don't use any additional parallel or monolingual corpora except for the datasets provided by the organizers. For both Czech-Polish and Spanish-Portuguese, we used all available parallel training data, which in the case of Czech-Polish consisted of about 2.2 million sentences and about 4.5 million sentences in the case of Spanish-Portuguese. Also, we used all the target-side monolingual data, which was 1.2 million sentences for Polish and 10.9 million sentences for Portuguese.

\paragraph{Preprocessing}
Our NMT model was trained on a combination of the original Czech-Polish parallel corpus together with pseudo-parallel corpus obtained from translating Polish monolingual data to Czech with Moses. Additionally, the development corpus was split into two sets: first containing 2k sentences and second containing 1k sentences, where the former was added to the training data and the latter was used for validation purposes.

Our Phrase-based model was trained on a combination of the original Spanish-Portuguese parallel corpus together with 2k sentences from the dev corpus. Specifically, the development corpus was split into two sets: first containing 2k sentences and second containing 1k sentences, where the former was added to the training data and the latter was used for validation purposes.

Then we followed the standard preprocessing scheme, where training, dev and test data are normalized, tokenized and truecased using \textit{Moses}\footnote{https://github.com/moses-smt/mosesdecoder} scripts. Additionally, training data was also cleaned with \texttt{clean-corpus-n.perl} script from \textit{Moses}. Finally, to allow open-vocabulary, we learned and applied byte-pair encoding (BPE)\footnote{https://github.com/rsennrich/subword-nmt} for the concatenation of the source and target languages with 16k operations.
The postprocessing was done in reverse order and included detruecasing and detokenization. 

%% file: experimental.tex
\subsection{Parameter Details}
\paragraph{Phrase-based}
For the Phrase-based systems we used Moses \cite{koehn2007moses}, which is a statistical machine translation system.
In order to build our model, we used generally the default parameters which include: grow-diagonal-final-and word alignment, lexical msd-bidirectional-fe reordering model trained, lexical weights, binarized and compacted phrase table with 4 score components and 4 threads used for conversion, 5-gram, binarized, loading-on-demand language model with Kneser-Ney smoothing and trie data structure without pruning; and MERT (Minimum Error Rate Training) optimisation with 100 n-best list generated and 16 threads.

\paragraph{Neural-based}
Our neural network model is based on the Transformer architecture (as described in section \ref{sec:neural_approach}) implemented by Facebook in the \textit{fairseq} toolkit\footnote{https://github.com/pytorch/fairseq}. The following hyperparameter configuration was used: 6 attention layers in the encoder and the decoder, with 4 attention heads per layer, embedding dimension of 512, maximum number of tokens per batch set to 4000, Adam optimizer with $\beta_1$ = 0.90, $\beta_2$ = 0.98, varied learning rate with the inverse square root of the step number (warmup steps equal 4000), dropout regularization and label smoothing set to 0.1, weight decay and gradient clipping threshold set to 0.

\paragraph{System Combination} The key parameter in the system combination with back-translation, explained in section \ref{sec:backtranslation}, is the score. Hence, we used the BLEU score \cite{papineni2002bleu} at the sentence level, implemented as \textit{sentence-bleu} in \textit{Moses}. Furthermore, we assigned equal weights to both Phrase and Neural-based translations and back-translations.

As contrastive approaches for system combination, we used two additional strategies:  Minimum Bayes Risk \cite{mbr} and the length ratio. In the former case, we used the implementation available in \textit{Moses}. In the latter approach, the ratio was computed as the number of words in the translation divided by the number of words in the source input. Sentence translations that gave a length ratio closer to 1 were selected. In the case of ties, we kept the sentence from the system that scored the best according to Table \ref{results}.

%% file: results.tex

The results provided in Table \ref{results-pb-nmt} show BLEU scores for the direct Phrase-based and Neural-based MT systems. Also, we report on experiments with incorporating monolingual data in two ways: either using a monolingual corpus both on the source and target sides (\textit{monolingual}) or using the back-translation system to produce a translation of a monolingual corpus (\textit{pseudo corpus}). Interestingly, we observe that the \textit{monolingual} approach harms the performance of the system even in the case of similar languages.

\begin{table}[!h]
\small
\centering
\caption{\label{results-pb-nmt}Phrase-based (PB) and Neural-based (NMT) results.}
\begin{tabular}{|l|c|c|}
\hline
& \multicolumn{1}{c|}{\textbf{CS-PL}} & \multicolumn{1}{c|}{\textbf{ES-PT}} \\ \hline

\multicolumn{1}{|l|}{PB}        
& 9.87 & 64.96 \\ \hline
\multicolumn{1}{|l|}{NMT}
& 11.69 & 58.40 \\ \hline
\multicolumn{1}{|l|}{NMT + monolingual}
& 10.91 & 52.37 \\ \hline
\multicolumn{1}{|l|}{NMT + pseudo corpus}       & 12.76 & 59.47 \\ \hline
\end{tabular}
\vspace{-1em}
\end{table}


\begin{table}[!h]
\small
\centering
\caption{\label{backtranslationres} Back-translation system results.}
\begin{tabular}{|l|l|c|c|}
\hline
1st system & 2nd system & \multicolumn{1}{c|}{\textbf{PL-CS}} & \multicolumn{1}{c|}{\textbf{PT-ES}} \\ \hline
\multicolumn{1}{|l|}{\multirow{2}{*}{PB}} & \multicolumn{1}{|l|}{PB}       & 44.34   &   84.62        \\ \cline{2-4}
\multicolumn{1}{|l|}{} & \multicolumn{1}{|l|}{NMT}        &  24.51  & 66.15          \\ \hline
\multicolumn{1}{|l|}{\multirow{2}{*}{NMT}} & \multicolumn{1}{|l|}{PB}    &            32.47 & 72.02       \\ \cline{2-4}
\multicolumn{1}{|l|}{} & \multicolumn{1}{|l|}{NMT}    &            27.31 & 59.56       \\ \hline
\end{tabular}
\vspace{-1em}
\end{table}

\begin{table}[!h]
\small
\centering
\caption{\label{resultssyscomb}System Combination results.}
\label{results}
\begin{tabular}{|l|c|c|}
\hline
& \textbf{CS-PL} & \textbf{ES-PT} \\ \hline
\multicolumn{1}{|l|}{MBR}        &     12.75      & 63.03          \\ \hline
\multicolumn{1}{|l|}{Back-translation}    & 10.71           &    63.71        \\ \hline
\end{tabular}
\vspace{-1em}
\end{table}

As presented in Table \ref{resultssyscomb}, our proposed system combinations, employing either MBR or the back-translation approach, did not achieve any significant improvements. The MBR strategy was applied to all systems from Table 1, which means that for the Czech-Polish pair we used 4 systems and for Spanish-Portuguese we used 3 systems. Back-translation results were evaluated with respect to the systems in Table \ref{backtranslationres} and the system combination with back-translation was created using the best two systems from Table \ref{results-pb-nmt}. 

In order to analyze the reason behind the weak performance of the system combination with back-translation, we evaluated the correlation between the quality of each translated sentence (generated using PB and NMT systems) and the quality of back-translations (both for PB and NMT systems) on the validation set. For any combination, Czech-Polish or Spanish-Portuguese, correlation varies between 0.2 and 0.4, which explains the poor performance of back-translation as a quality estimation metric.

%% file: discussion.tex
Although Czech and Polish belong to the same family of languages (Slavic) and share the same subgroup (Western Slavic), the BLEU score obtained by our winning system is relatively low comparing to other pairs of similar languages (e.g. Spanish and Portuguese). It may seem surprising considering some common characteristics shared by both languages, such as 7 noun cases, 2 number cases, 3 noun gender cases as well as 3 tenses among others.

Low performance on this task could be explained by the language distance. Considering the metric proposed by \newcite{GAMALLO:2017}, which is based on perplexity as a distance measure between languages, the distance between Czech and Polish is 27 while for Spanish-Portuguese is 7. The very same metric used to evaluate the distance of Czech and Polish from other Slavic languages (i.e. Slovak and Russian) shows that Polish is the most distant language within this group (see Table \ref{tab:distmeas}). In general, distances between Latin languages are smaller than between Slavic ones.

\begin{table}[!h]
\small
\centering
\caption{\label{tab:distmeas}Distances between Slavic and Latin languages. Examples across families.}
\begin{tabular}{|c|c|c|c|c|c|}
\hline
\multicolumn{2}{|c|}{Slavic} & \multicolumn{2}{c|}{Latin} & \multicolumn{2}{c|}{Mix} \\ \hline
pair & dist. & pair & dist. & pair & dist.\\ \hline
CS-PL & 27 & ES-PT & 7 & ES-CS & 37\\ 
CS-SL & 8 & ES-FR & 15 & ES-PL & 44\\
CS-RU & 21 & ES-RO & 20& PT-CS & 31\\
PL-SL &  24 & PT-FR & 15& PT-PL & 38\\
PL-RU & 34 & PT-RO & 22& & \\
\hline
\end{tabular}
\vspace{-1em}
\end{table}

While  Czech and Polish languages are highly inflected, which poses a challenge, we hypothesize that one of the reasons for the low BLEU score lies also in the difference of the alphabets. Even though both alphabets are based on the Latin script, they include letters with diacritics -- \textit{ą, ć, ę, ł, ń, ó, ś, ź, ż} in Polish, and \textit{á, č, ď, é, ě, ch, í, ň, ó, ř, š, ť, ú, ů, ý, ž} in Czech.
The total number of unique letters in Polish is 32, while in the Czech language there are 42 letters.
Moreover, some letters are used only in the case of foreign words, such as \textit{q}, \textit{x} (in Czech and Polish), \textit{w} (in Czech), and \textit{v} (in Polish).

%% file: futurework.tex
In the future we plan to extend our research in the following directions. First, we would like to explore how removing diacritics on the source-side would impact the performance of our system for the Czech-Polish language pair. Furthermore, we would like to study the performance of our system combination while applying various quality estimation approaches. We would be interested in experimenting with the reward score introduced by \newcite{he2016dual}, which is a linear combination of language model score and the reconstruction probability of the back-translated sentence, as well as with other quality measures implemented in the \textit{OpenKiwi} \cite{kepler2019openkiwi} toolkit\footnote{https://github.com/Unbabel/OpenKiwi}.

%% file: acl2019.bbl
\begin{thebibliography}{26}
\expandafter\ifx\csname natexlab\endcsname\relax\def\natexlab#1{#1}\fi

\bibitem[{Artetxe et~al.(2018)Artetxe, Labaka, and
  Agirre}]{artetxe2018unsupervised-smt}
Mikel Artetxe, Gorka Labaka, and Eneko Agirre. 2018.
\newblock \href {https://arxiv.org/abs/1809.01272} {Unsupervised statistical
  machine translation}.
\newblock In \emph{Proceedings of the 2018 Conference on Empirical Methods in
  Natural Language Processing}, Brussels, Belgium. Association for
  Computational Linguistics.

\bibitem[{Bahdanau et~al.(2015)Bahdanau, Cho, and Bengio}]{bahdanau2014neural}
Dzmitry Bahdanau, Kyunghyun Cho, and Yoshua Bengio. 2015.
\newblock \href {http://arxiv.org/abs/1409.0473} {Neural machine translation by
  jointly learning to align and translate}.
\newblock In \emph{3rd International Conference on Learning Representations,
  {ICLR} 2015, San Diego, CA, USA, May 7-9, 2015, Conference Track
  Proceedings}.

\bibitem[{Barrault et~al.(2019)Barrault, Bojar, Costa-juss{\`a}, Federmann,
  Fishel, Graham, Haddow, Huck, Koehn, Malmasi, Monz, M{\"u}ller, Pal, Post,
  and Zampieri}]{barrault-EtAl:2019:WMT}
Lo{\"i}c Barrault, Ond\v{r}ej Bojar, Marta~R. Costa-juss{\`a}, Christian
  Federmann, Mark Fishel, Yvette Graham, Barry Haddow, Matthias Huck, Philipp
  Koehn, Shervin Malmasi, Christof Monz, Mathias M{\"u}ller, Santanu Pal, Matt
  Post, and Marcos Zampieri. 2019.
\newblock Findings of the 2019 conference on machine translation (wmt19).
\newblock In \emph{Proceedings of the Fourth Conference on Machine Translation,
  Volume 2: Shared Task Papers}, Florence, Italy. Association for Computational
  Linguistics.

\bibitem[{Cho et~al.(2014)Cho, van Merrienboer, G{\"{u}}l{\c{c}}ehre, Bahdanau,
  Bougares, Schwenk, and Bengio}]{cho2014learning}
Kyunghyun Cho, Bart van Merrienboer, {\c{C}}aglar G{\"{u}}l{\c{c}}ehre, Dzmitry
  Bahdanau, Fethi Bougares, Holger Schwenk, and Yoshua Bengio. 2014.
\newblock \href {http://aclweb.org/anthology/D/D14/D14-1179.pdf} {Learning
  phrase representations using {RNN} encoder-decoder for statistical machine
  translation}.
\newblock In \emph{Proceedings of the 2014 Conference on Empirical Methods in
  Natural Language Processing, {EMNLP} 2014, October 25-29, 2014, Doha, Qatar},
  pages 1724--1734.

\bibitem[{Currey et~al.(2017)Currey, Miceli~Barone, and
  Heafield}]{currey-etal-2017-copied}
Anna Currey, Antonio~Valerio Miceli~Barone, and Kenneth Heafield. 2017.
\newblock \href {https://doi.org/10.18653/v1/W17-4715} {Copied monolingual data
  improves low-resource neural machine translation}.
\newblock In \emph{Proceedings of the Second Conference on Machine
  Translation}, pages 148--156, Copenhagen, Denmark. Association for
  Computational Linguistics.

\bibitem[{Gamallo et~al.(2017)Gamallo, Pichel, and Alegria}]{GAMALLO:2017}
Pablo Gamallo, José~Ramom Pichel, and Iñaki Alegria. 2017.
\newblock \href {https://gramatica.usc.es/~gamallo/artigos-web/PHYSICA2017.pdf}
  {From language identification to language distance}.
\newblock \emph{Physica A: Statistical Mechanics and its Applications}, 484:152
  -- 162.

\bibitem[{Gehring et~al.(2017)Gehring, Auli, Grangier, Yarats, and
  Dauphin}]{gehring2017}
Jonas Gehring, Michael Auli, David Grangier, Denis Yarats, and Yann~N. Dauphin.
  2017.
\newblock \href {http://arxiv.org/abs/1705.03122} {Convolutional sequence to
  sequence learning}.
\newblock In \emph{Proceedings of the 34th International Conference on Machine
  Learning - Volume 70}, ICML'17, pages 1243--1252.

\bibitem[{He et~al.(2016{\natexlab{a}})He, Xia, Qin, Wang, Yu, Liu, and
  Ma}]{he2016dual}
Di~He, Yingce Xia, Tao Qin, Liwei Wang, Nenghai Yu, Tie-Yan Liu, and Wei-Ying
  Ma. 2016{\natexlab{a}}.
\newblock \href
  {https://papers.nips.cc/paper/6469-dual-learning-for-machine-translation.pdf}
  {Dual learning for machine translation}.
\newblock In \emph{Advances in Neural Information Processing Systems}, pages
  820--828.

\bibitem[{He et~al.(2016{\natexlab{b}})He, Zhang, Ren, and Sun}]{he2015}
Kaiming He, Xiangyu Zhang, Shaoqing Ren, and Jian Sun. 2016{\natexlab{b}}.
\newblock \href {http://arxiv.org/abs/1512.03385} {Deep residual learning for
  image recognition}.
\newblock In \emph{Proceedings of the IEEE conference on computer vision and
  pattern recognition}, pages 770--778.

\bibitem[{Hochreiter and Schmidhuber(1997)}]{hochreiter1997}
Sepp Hochreiter and J\"{u}rgen Schmidhuber. 1997.
\newblock \href {https://doi.org/10.1162/neco.1997.9.8.1735} {Long short-term
  memory}.
\newblock \emph{Neural Comput.}, 9(8):1735--1780.

\bibitem[{Kalchbrenner and Blunsom(2013)}]{kalchbrenner2013}
Nal Kalchbrenner and Phil Blunsom. 2013.
\newblock \href {https://www.aclweb.org/anthology/D13-1176} {Recurrent
  continuous translation models}.
\newblock In \emph{Proceedings of the 2013 Conference on Empirical Methods in
  Natural Language Processing}, pages 1700--1709. Association for Computational
  Linguistics.

\bibitem[{Kalchbrenner et~al.(2016)Kalchbrenner, Espeholt, Simonyan, van~den
  Oord, Graves, and Kavukcuoglu}]{kalchbrenner2O16}
Nal Kalchbrenner, Lasse Espeholt, Karen Simonyan, A{\"{a}}ron van~den Oord,
  Alex Graves, and Koray Kavukcuoglu. 2016.
\newblock \href {http://arxiv.org/abs/1610.10099} {Neural machine translation
  in linear time}.
\newblock \emph{CoRR}, abs/1610.10099.

\bibitem[{Kepler et~al.(2019)Kepler, Tr{\'e}nous, Treviso, Vera, and
  Martins}]{kepler2019openkiwi}
F{\'a}bio Kepler, Jonay Tr{\'e}nous, Marcos Treviso, Miguel Vera, and
  Andr{\'e}~FT Martins. 2019.
\newblock \href {https://arxiv.org/abs/1902.08646} {Openkiwi: An open source
  framework for quality estimation}.
\newblock \emph{arXiv preprint arXiv:1902.08646}.

\bibitem[{Koehn et~al.(2007)Koehn, Hoang, Birch, Callison-Burch, Federico,
  Bertoldi, Cowan, Shen, Moran, Zens et~al.}]{koehn2007moses}
Philipp Koehn, Hieu Hoang, Alexandra Birch, Chris Callison-Burch, Marcello
  Federico, Nicola Bertoldi, Brooke Cowan, Wade Shen, Christine Moran, Richard
  Zens, et~al. 2007.
\newblock \href {https://aclweb.org/anthology/papers/P/P07/P07-2045/} {Moses:
  Open source toolkit for statistical machine translation}.
\newblock In \emph{Proceedings of the 45th annual meeting of the association
  for computational linguistics companion volume proceedings of the demo and
  poster sessions}, pages 177--180.

\bibitem[{Koehn et~al.(2003)Koehn, Och, and Marcu}]{koehn:2003}
Philipp Koehn, Franz~J. Och, and Daniel Marcu. 2003.
\newblock \href {https://www.aclweb.org/anthology/N03-1017} {Statistical
  phrase-based translation}.
\newblock In \emph{Proceedings of the 2003 Human Language Technology Conference
  of the North {A}merican Chapter of the Association for Computational
  Linguistics}, pages 127--133.

\bibitem[{Kumar and Byrne(2004)}]{mbr}
Shankar Kumar and William Byrne. 2004.
\newblock \href {https://www.aclweb.org/anthology/N04-1022} {Minimum
  {B}ayes-risk decoding for statistical machine translation}.
\newblock In \emph{HLT-NAACL 2004: Main Proceedings}, pages 169--176, Boston,
  Massachusetts, USA. Association for Computational Linguistics.

\bibitem[{Lample et~al.(2018)Lample, Ott, Conneau, Denoyer, and
  Ranzato}]{lample-etal-2018-phrase}
Guillaume Lample, Myle Ott, Alexis Conneau, Ludovic Denoyer, and Marc{'}Aurelio
  Ranzato. 2018.
\newblock \href {https://www.aclweb.org/anthology/D18-1549} {Phrase-based {\&}
  neural unsupervised machine translation}.
\newblock In \emph{Proceedings of the 2018 Conference on Empirical Methods in
  Natural Language Processing}, pages 5039--5049, Brussels, Belgium.
  Association for Computational Linguistics.

\bibitem[{Luong et~al.(2015)Luong, Pham, and Manning}]{luong2015}
Thang Luong, Hieu Pham, and Christopher~D. Manning. 2015.
\newblock \href {https://doi.org/10.18653/v1/D15-1166} {Effective approaches to
  attention-based neural machine translation}.
\newblock In \emph{Proceedings of the 2015 Conference on Empirical Methods in
  Natural Language Processing}, pages 1412--1421, Lisbon, Portugal. Association
  for Computational Linguistics.

\bibitem[{Marie and Fujita(2018)}]{marie-fujita-2018-smorgasbord}
Benjamin Marie and Atsushi Fujita. 2018.
\newblock \href {https://www.aclweb.org/anthology/W18-1811} {A smorgasbord of
  features to combine phrase-based and neural machine translation}.
\newblock In \emph{Proceedings of the 13th Conference of the Association for
  Machine Translation in the {A}mericas (Volume 1: Research Papers)}, pages
  111--124, Boston, MA. Association for Machine Translation in the Americas.

\bibitem[{Papineni et~al.(2002)Papineni, Roukos, Ward, and
  Zhu}]{papineni2002bleu}
Kishore Papineni, Salim Roukos, Todd Ward, and Wei-Jing Zhu. 2002.
\newblock \href {https://www.aclweb.org/anthology/P02-1040.pdf} {Bleu: a method
  for automatic evaluation of machine translation}.
\newblock In \emph{Proceedings of the 40th annual meeting on association for
  computational linguistics}, pages 311--318. Association for Computational
  Linguistics.

\bibitem[{Sennrich et~al.(2016{\natexlab{a}})Sennrich, Haddow, and
  Birch}]{sennrich2016wmt}
Rico Sennrich, Barry Haddow, and Alexandra Birch. 2016{\natexlab{a}}.
\newblock \href {https://doi.org/10.18653/v1/W16-2323} {{E}dinburgh neural
  machine translation systems for {WMT} 16}.
\newblock In \emph{Proceedings of the First Conference on Machine Translation},
  pages 371--376, Berlin, Germany. Association for Computational Linguistics.

\bibitem[{Sennrich et~al.(2016{\natexlab{b}})Sennrich, Haddow, and
  Birch}]{sennrich:2016}
Rico Sennrich, Barry Haddow, and Alexandra Birch. 2016{\natexlab{b}}.
\newblock \href {https://doi.org/10.18653/v1/P16-1009} {Improving neural
  machine translation models with monolingual data}.
\newblock In \emph{Proceedings of the 54th Annual Meeting of the Association
  for Computational Linguistics (Volume 1: Long Papers)}, pages 86--96, Berlin,
  Germany. Association for Computational Linguistics.

\bibitem[{Sutskever et~al.(2014)Sutskever, Vinyals, and Le}]{sutskever2014}
Ilya Sutskever, Oriol Vinyals, and Quoc~V Le. 2014.
\newblock \href
  {http://papers.nips.cc/paper/5346-sequence-to-sequence-learning-with-neural-networks.pdf}
  {Sequence to sequence learning with neural networks}.
\newblock In Z.~Ghahramani, M.~Welling, C.~Cortes, N.~D. Lawrence, and K.~Q.
  Weinberger, editors, \emph{Advances in Neural Information Processing Systems
  27}, pages 3104--3112. Curran Associates, Inc.

\bibitem[{Vaswani et~al.(2017)Vaswani, Shazeer, Parmar, Uszkoreit, Jones,
  Gomez, Kaiser, and Polosukhin}]{vaswani2017attention}
Ashish Vaswani, Noam Shazeer, Niki Parmar, Jakob Uszkoreit, Llion Jones,
  Aidan~N Gomez, {\L}ukasz Kaiser, and Illia Polosukhin. 2017.
\newblock \href
  {https://papers.nips.cc/paper/7181-attention-is-all-you-need.pdf} {Attention
  is all you need}.
\newblock In \emph{Advances in Neural Information Processing Systems}, pages
  5998--6008.

\bibitem[{Wu et~al.(2016)Wu, Schuster, Chen, Le, Norouzi, Macherey, Krikun,
  Cao, Gao, Macherey, Klingner, Shah, Johnson, Liu, Kaiser, Gouws, Kato, Kudo,
  Kazawa, Stevens, Kurian, Patil, Wang, Young, Smith, Riesa, Rudnick, Vinyals,
  Corrado, Hughes, and Dean}]{wu2016}
Yonghui Wu, Mike Schuster, Zhifeng Chen, Quoc~V. Le, Mohammad Norouzi, Wolfgang
  Macherey, Maxim Krikun, Yuan Cao, Qin Gao, Klaus Macherey, Jeff Klingner,
  Apurva Shah, Melvin Johnson, Xiaobing Liu, Lukasz Kaiser, Stephan Gouws,
  Yoshikiyo Kato, Taku Kudo, Hideto Kazawa, Keith Stevens, George Kurian,
  Nishant Patil, Wei Wang, Cliff Young, Jason Smith, Jason Riesa, Alex Rudnick,
  Oriol Vinyals, Greg Corrado, Macduff Hughes, and Jeffrey Dean. 2016.
\newblock \href {http://arxiv.org/abs/1609.08144} {Google's neural machine
  translation system: Bridging the gap between human and machine translation}.
\newblock \emph{CoRR}, abs/1609.08144.

\bibitem[{Zhou et~al.(2016)Zhou, Cao, Wang, Li, and Xu}]{zhou2016}
Jie Zhou, Ying Cao, Xuguang Wang, Peng Li, and Wei Xu. 2016.
\newblock \href {https://arxiv.org/pdf/1606.04199} {Deep recurrent models with
  fast-forward connections for neural machine translation.}
\newblock \emph{TACL}, 4:371--383.

\end{thebibliography}
